\documentclass{article}

\usepackage[final]{neurips_2026}

\makeatletter
\renewcommand{\@noticestring}{}
\makeatother

\usepackage[utf8]{inputenc}
\usepackage[T1]{fontenc}
\usepackage{hyperref}
\usepackage{url}
\usepackage{booktabs}
\usepackage{amsfonts}
\usepackage{amsmath,amssymb}
\usepackage{nicefrac}
\usepackage{microtype}
\usepackage{xcolor}
\usepackage{graphicx}
\usepackage{multirow}
\usepackage{algorithm}
\usepackage{algorithmicx}
\usepackage{algpseudocode}
\usepackage{siunitx}
\usepackage{makecell}
\usepackage{enumitem}
\usepackage[capitalize,noabbrev]{cleveref}

\title{SmartRAG: Native Graph-Based RAG for Mobile Device}

\author{%
  Zhihan Jiang\textsuperscript{1}, Meng Li\textsuperscript{1}, Shenghao Liu\textsuperscript{2}, Keran Li\textsuperscript{1}, Ruiben Zhou\textsuperscript{1} \And
  Wei Wang\textsuperscript{1}, Xianjun Deng\textsuperscript{2}, Shuai Wang\textsuperscript{3}, Haipeng Dai\textsuperscript{1} \\[6pt]
  \textsuperscript{1}Nanjing University \quad
  \textsuperscript{2}HUST \quad
  \textsuperscript{3}Southeast University \\[6pt]
  \texttt{zhihan\_jiang@smail.nju.edu.cn}, \texttt{meng@nju.edu.cn}
}

\begin{document}

\maketitle

\begin{abstract}
Deploying large language models (LLMs) as personal assistants on mobile devices demands privacy, low latency, and offline availability, yet the computational cost of giant models clashes with strict edge-hardware budgets. We argue that this tension cannot be resolved by model compression alone; it requires decomposing on-device intelligence into complementary functional roles. We present \textbf{SmartRAG}, a fully on-device framework that organizes an intelligent assistant around four coordinated modules---\emph{Perception}, \emph{Memory}, \emph{Focus}, and \emph{Thinking}. At the core of SmartRAG is \textbf{EvoNER}, a continually learnable named-entity recognizer that incrementally expands its label inventory through teacher-distilled updates, enabling the system to absorb previously unseen entity types without retraining the backbone LLM. Extracted knowledge is stored in \textbf{MRGraph}, a three-layer provenance-preserving knowledge graph, and retrieved at query time through a hybrid pipeline combining graph traversal, lexical matching, and dense semantic search. The on-device LLM is invoked only for high-value semantic operations---labeling, planning, and answer synthesis---keeping inference costs bounded. Experiments on four QA benchmarks (TriviaQA, Natural Questions, HotpotQA, MultiHopQA) show that SmartRAG with a quantized 1.7B-parameter backbone achieves multi-hop reasoning performance competitive with models up to 18$\times$ larger, while running entirely on commodity smartphones within practical memory and latency envelopes.
\end{abstract}

% ======================================================================
\section{Introduction}
\label{sec:intro}
% ======================================================================
In recent years, as the application of large language models (LLMs) in real-world interactive tasks continues to deepen, memory is gradually emerging as a critical capability in LLM system determining whether the system can support long-term personalization, continuous knowledge updates, and contextually consistent reasoning. This issue is of particular significance for on-device scenarios, as the core value of such deployments lies precisely in data locality, privacy protection, and personalized services capabilities that inherently require the system to establish and maintain a local memory built around the user's private data, device-local content, and environmental context. Given the strict constraints on system-memory (i.e., RAM usage and runtime memory footprint on edge devices), power consumption, and latency budgets inherent to devices, heavyweight LLMs are difficult to deploy directly; furthermore, relying solely on the limited parametric memory of lightweight models often proves insufficient to support continuous updates, user customization, and knowledge-intensive reasoning. Retrieval-Augmented Generation (RAG) addresses this challenge by integrating LLMs with external, non-parametric memory, thereby providing a memory mechanism that is updatable, structured, and more easily personalized; consequently, it is widely regarded as a highly promising solution.

Although on-device RAG has recently achieved preliminary progress, most existing methods remain centered on naive vector retrieval—that is, relying on semantic similarity between queries and document snippets to facilitate memory access. While this paradigm is well-suited for handling problems involving local matching, it struggles to address complex queries that require cross-segment evidence aggregation, entity linking, and multi-step reasoning. To address this limitation, on-cloud RAG research has begun to leverage graph structures to enhance the retrieval process, explicitly modeling the connections between entities, relations, and evidence to improve multi-hop question-answering capabilities. However, the majority of existing graph-based RAG methods are built upon powerful models, relying heavily on models with tens of billions of parameters to execute critical steps such as graph construction, summarization, and relation organization. While this premise may hold true in cloud-based environments, it clearly fails in on-device scenarios: on-device systems simply cannot bear the computational, latency, energy, and cost overheads associated with models of such scale.

Based on this issue, we believe that the key to on-device RAG lies not in directly compressing cloud-based graph RAGs onto the device, but in rethinking how structured memory should be built and utilized under resource-constrained conditions. In other words, graph construction capabilities do not necessarily stem from the generation capabilities of super-powerful models; they can also arise from more sophisticated system architecture design. By rationally dividing module responsibilities, controlling the granularity of knowledge organization, and designing incrementally evolving memory formation mechanisms, lightweight components can also gradually build a structured knowledge foundation that supports complex reasoning. Unlike traditional graph-based RAG methods, our approach emphasizes a systematic decoupling of knowledge extraction, memory organization, and reasoning support, allowing these processes to be collaboratively completed by multiple lightweight modules under on-device constraints. This provides an implementation path for on-device RAGs that differs from the cloud paradigm.

However, several key technical challenges remain to be addressed before this approach can be truly implemented on device. First, due to budget constraints related to system-memory, power consumption, and latency, only lightweight models are deployable on device. These models are insufficient in open environments to identify unknown entities, long-tail concepts, and new knowledge, making it difficult to directly construct high-quality structured knowledge like large cloud-based models. Therefore, the system must possess the ability to discover, aggregate, and continuously absorb unseen entities. Furthermore, complex queries often require entity association tracing and multi-step retrieval across multiple pieces of evidence. However, edge scenarios cannot rely on powerful models to repeatedly perform query decomposition, path planning, and graph exploration. This makes achieving structured focus for complex problems within a limited budget another core challenge. Finally, even after relevant evidence has been retrieved, lightweight generation models struggle to reliably complete reasoning within lengthy and noisy contexts. Therefore, a controlled evidence organization method is needed to confine the final generation to a compact and high-value context.

Based on this understanding, we propose \textbf{SmartRAG}, a structured RAG framework for on-device scenarios. Its core objective is to build external memory capable of supporting complex reasoning under lightweight models and limited resources. Unlike existing methods that rely on powerful models to directly complete graph construction and graph-based reasoning, SmartRAG decouples knowledge extraction, memory organization, retrieval focus, and answer generation through modular design, enabling these processes to operate collaboratively under edge-side constraints. Thus, SmartRAG transforms the structured knowledge capability from LLMs capability into a system architecture problem, providing a more practical implementation method for on-device RAGs.

Our contributions are as follows:
\begin{itemize}[leftmargin=*,itemsep=2pt,topsep=2pt]
    \item We identify the core feasibility problem of on-device structured RAG: complex reasoning requires structured and updatable memory, but existing graph-construction pipelines depend too heavily on powerful LLMs to remain sustainable on edge devices.
    \item We present \textbf{SmartRAG}, an on-device structured RAG method that reallocates high-frequency memory writing to lightweight trainable components while reserving the LLM for sparse, high-value semantic operations.
    \item We develop \textbf{EvoNER}, a continually learnable entity recognizer that absorbs previously unseen semantic categories through lightweight incremental updates, enabling the system to recognize and organize emerging concepts without retraining the backbone LLM.
    \item We integrate provenance-preserving structured memory and hybrid retrieval into a full on-device pipeline, and demonstrate on real smartphones that this design yields strong end-to-end QA performance under strict device constraints.
\end{itemize}

% ======================================================================
\section{Related work}
\label{sec:related}
% ======================================================================

\paragraph{On-device LLMs.}
Recent advances in quantization~\citep{frantar2023gptq} and efficient inference engines~\citep{llamacpp2023} have made it feasible to run billion-parameter models on mobile hardware~\citep{xu2024device}. Models such as Qwen3~\citep{qwen3report} and Llama~3~\citep{llama3herd2024} offer compact variants targeting edge deployment. However, these efforts focus on the \emph{inference engine} and do not address the \emph{memory} and \emph{continuous adaptation} requirements of a personal assistant.

\paragraph{Retrieval-augmented generation.}
RAG~\citep{lewis2020retrieval} augments LLMs with external knowledge at inference time. Dense passage retrieval~\citep{karpukhin2020dense} provides the standard retrieval backbone. Self-RAG~\citep{asai2023self} internalizes retrieval behaviors through specialized training, and xRAG~\citep{cheng2024xrag} compresses retrieved context into compact representations. Graph-based approaches~\citep{edge2024local} organize retrieved knowledge into structured graphs for improved multi-hop reasoning. Multi-hop QA methods~\citep{trivedi2023interleaving,press2023measuring} interleave retrieval with chain-of-thought reasoning. Most of these systems assume cloud-scale compute; our work targets their adaptation to strict on-device budgets.

\paragraph{Named entity recognition and continual learning.}
Span-based NER~\citep{yu2020named,fu2021spanner} formulates entity detection as span classification over enumerated candidates, avoiding the label dependency of sequential taggers. Continual learning for NER~\citep{monaikul2021continual} addresses the challenge of absorbing new entity types without forgetting old ones, commonly using knowledge distillation~\citep{hinton2015distilling} or regularization~\citep{kirkpatrick2017overcoming}. Our EvoNER combines span-based detection with a reserved-label classifier and teacher-distilled incremental updates, specifically designed for on-device expansion of the entity vocabulary.

\paragraph{Knowledge graphs for LLMs.}
The integration of knowledge graphs with LLMs~\citep{ji2021survey,pan2024unifying} has attracted growing attention. LLM-driven KG construction~\citep{edge2024local} achieves high-quality extraction but requires large models. Our approach separates the extraction pipeline into a lightweight, trainable perception layer and a sparse LLM control plane, achieving predictable ingestion cost on mobile devices.

% ======================================================================
\section{Method}
\label{sec:method}
% ======================================================================

\subsection{Architecture overview}
\label{sec:overview}

\begin{figure}[t]
\centering
\includegraphics[width=1\textwidth]{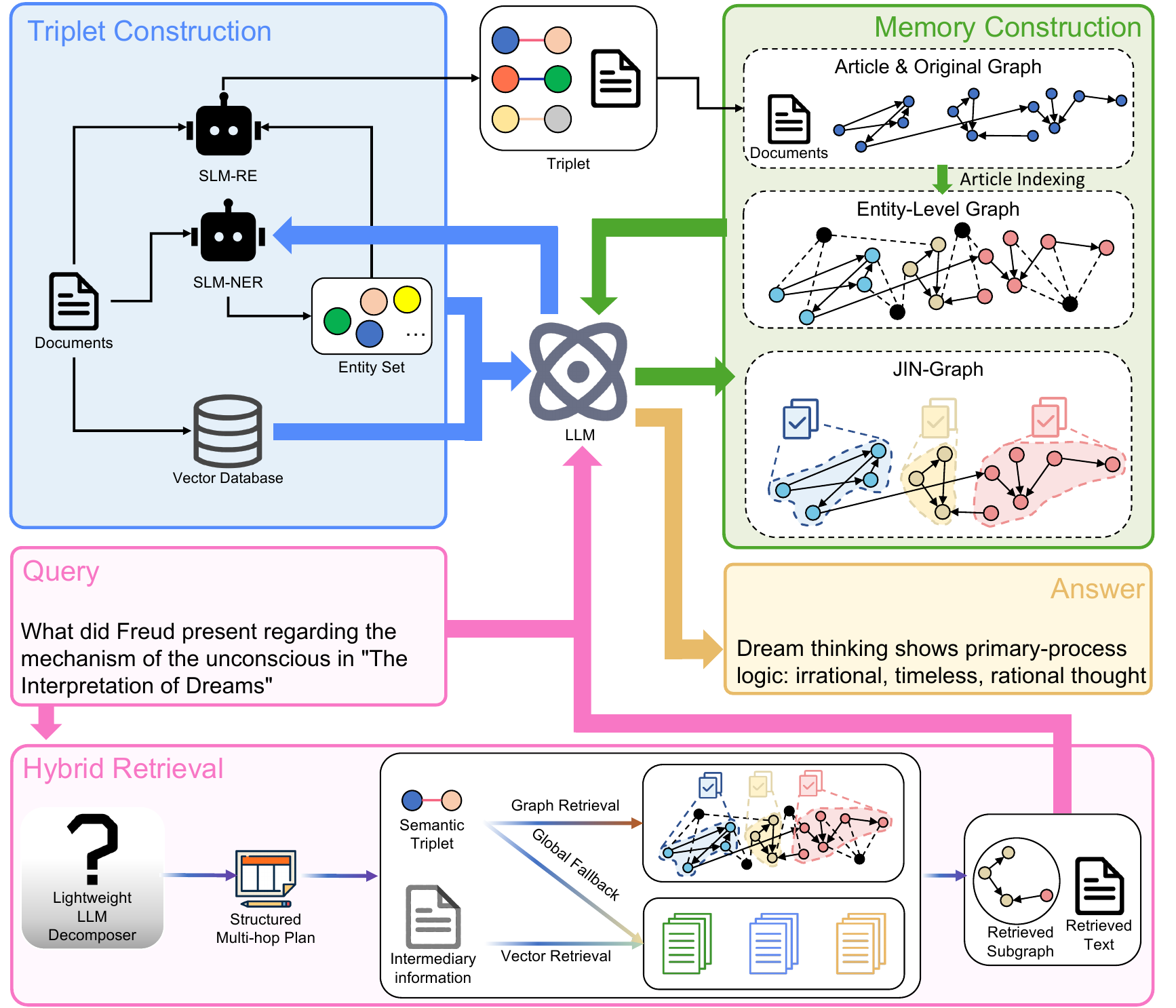}
\caption{Overview of the SmartRAG architecture. Incoming text flows through \emph{Perception} (EvoNER + relation extraction) into \emph{Memory} (MRGraph), which is queried by \emph{Focus} (hybrid retrieval) and consumed by \emph{Thinking} (LLM-driven planning and answer generation). Dashed arrows indicate asynchronous background processes (clustering, incremental training).}
\label{fig:system}
\end{figure}

The SmartRAG architecture (\cref{fig:system}) consists of four interacting modules. The \textbf{Perception} module transforms unstructured text into structured knowledge via EvoNER (an on-device NER component, \cref{sec:evoner}) and a lightweight relation classifier, producing entity--relation triplets. The \textbf{Memory} module organizes these triplets into MRGraph, a persistent knowledge structure that binds entities and relations to their source paragraphs for provenance (\cref{sec:memory}). At query time, the \textbf{Focus} module retrieves and ranks relevant evidence through a hybrid pipeline combining graph traversal, lexical matching, and dense semantic search (\cref{sec:focus}). The \textbf{Thinking} module invokes the on-device LLM only for high-value semantic operations: labeling emerging entity types, planning multi-hop retrieval, and synthesizing final answers (\cref{sec:thinking}).

\subsection{Perception: EvoNER and entity discovery}
\label{sec:evoner}

\paragraph{Span-based NER with character injection.}
We design \textbf{EvoNER} (\textbf{Evo}lutionary \textbf{N}amed \textbf{E}ntity \textbf{R}ecognition) to satisfy four on-device requirements: lightweight co-existence with the LLM, accurate entity boundary and type prediction, robustness to out-of-vocabulary surface forms, and support for continual label expansion. Given an input sentence, we encode it into sub-word representations and enumerate candidate spans from word-start positions up to a maximum width $W$:
\begin{equation}\label{eq:span}
    r=(i,j), \qquad 1 \le i \le j \le n,\qquad (j-i+1)\le W,
\end{equation}
where $r$ denotes a candidate span, $i$ and $j$ are its inclusive word-level start and end indices, $n$ is the sentence length in words, and $W$ caps span width ($W{=}10$ in our implementation), yielding $O(nW)$ candidates instead of all $O(n^2)$ spans. To improve robustness to misspellings and rare surface forms, we augment the contextual representation with a character-level feature injected at the first subword of each word:
\begin{equation}\label{eq:char}
    \mathbf{h}_t = [\mathbf{e}_t \,;\, m_t \cdot \mathbf{c}(t)],
\end{equation}
where $\mathbf{h}_t$ is the final representation at subword position $t$, $\mathbf{e}_t$ is the Transformer contextual embedding at $t$, $m_t\in\{0,1\}$ is $1$ on the first subword of a word and $0$ otherwise, $\mathbf{c}(t)$ is the character-level embedding aligned with that word, and $[\,;\,]$ denotes concatenation. Each candidate span is represented using boundary features, element-wise interaction, and a learned width embedding, then passed to a classifier.

\paragraph{Reserved-label mechanism for continual expansion.}
A key requirement for continual on-device learning is adding new entity types without reinitializing the classifier. We pre-allocate a fixed label capacity and partition the span classifier into a \emph{base} head (no-entity label plus deployed entity types) and \emph{reserved} rows for future types. For span $(i,j)$ with pooled representation $\mathbf{z}_{ij}\in\mathbb{R}^{d}$, logits decompose as
\begin{equation}
\begin{bmatrix}
    \boldsymbol{\ell}^{\text{base}}_{ij}\\
    \boldsymbol{\ell}^{\text{resv}}_{ij}
\end{bmatrix}
=
\begin{bmatrix}
    \mathbf{W}_{\text{base}}\\
    \mathbf{W}_{\text{resv}}
\end{bmatrix}
\mathbf{z}_{ij}
+
\begin{bmatrix}
    \mathbf{b}_{\text{base}}\\
    \mathbf{b}_{\text{resv}}
\end{bmatrix},
\qquad
\boldsymbol{\ell}_{ij} = [\boldsymbol{\ell}^{\text{base}}_{ij}\,;\,\boldsymbol{\ell}^{\text{resv}}_{ij}],
\end{equation}
where $\mathbf{W}_{\text{base}},\mathbf{b}_{\text{base}}$ (resp.\ $\mathbf{W}_{\text{resv}},\mathbf{b}_{\text{resv}}$) are base (resp.\ reserved) weights and biases. Reserved slots are initialized inactive ($\mathbf{b}_{\text{resv}} \ll 0$, $\mathbf{W}_{\text{resv}} \approx \mathbf{0}$). When new types are discovered, each is mapped to an unused reserved row without changing output dimensionality.

\paragraph{Incremental learning with teacher distillation.}
When new entity types emerge, EvoNER must incorporate the supervision without degrading performance on existing types. We pair the reserved-label design with teacher-guided distillation on the base-label subspace only. Let $\Omega(x)$ be the set of candidate spans in sentence $x$. The task loss matches offline training:
\begin{equation}
\mathcal{L}_{\mathrm{task}}
= \mathcal{L}_{\mathrm{span}}
+ \tfrac{1}{2}\big(\mathcal{L}_{\mathrm{start}} + \mathcal{L}_{\mathrm{end}}\big),
\end{equation}
where $\mathcal{L}_{\mathrm{span}}$ is masked span classification over candidates and $\mathcal{L}_{\mathrm{start}},\mathcal{L}_{\mathrm{end}}$ are auxiliary boundary losses on mention endpoints. The full incremental objective is
\begin{equation}
\mathcal{L}_{\mathrm{inc}} = \mathcal{L}_{\mathrm{task}} + \lambda_{\mathrm{KD}} \, \mathcal{L}_{\mathrm{KD}},
\end{equation}
with $\lambda_{\mathrm{KD}}$ trading stability against plasticity. For each span $r\in\Omega(x)$, let $\boldsymbol{\ell}_{r}^{T,\mathrm{base}},\boldsymbol{\ell}_{r}^{S,\mathrm{base}}\in\mathbb{R}^{1+n_{\mathrm{base}}}$ denote teacher and student logits \emph{restricted to the base subspace}, where $n_{\mathrm{base}}$ is the number of deployed base entity types (the leading dimension accounts for the no-entity logit). With temperature $T_{\mathrm{kd}}$ and confidence threshold $\tau$, define the gate
\begin{equation}
m_r = \mathbb{I}\!\left[
\max_{1\le k \le 1+n_{\mathrm{base}}} \big[\mathrm{softmax}\big(\boldsymbol{\ell}_{r}^{T,\mathrm{base}} / T_{\mathrm{kd}}\big)\big]_k \ge \tau
\right],
\end{equation}
i.e., the gate is on when the teacher's peak probability on base classes exceeds $\tau$.
The distillation loss is
\begin{equation}
\mathcal{L}_{\mathrm{KD}}
=
\frac{T_{\mathrm{kd}}^{2}}{\sum_{r\in\Omega(x)} m_r + \epsilon}
\sum_{r\in\Omega(x)} m_r \,
\mathrm{KL}\!\left(
\mathrm{softmax}\!\big(\boldsymbol{\ell}_{r}^{T,\mathrm{base}} / T_{\mathrm{kd}}\big)
\,\big\|\,
\mathrm{softmax}\!\big(\boldsymbol{\ell}_{r}^{S,\mathrm{base}} / T_{\mathrm{kd}}\big)
\right),
\end{equation}
where $\epsilon$ avoids division by zero. Reserved logits are \emph{not} distilled, so new types remain plastic while the teacher anchors old boundaries. The teacher is frozen for an entire incremental round, then the updated student may become the next teacher.

\paragraph{Three-stage parameter-efficient updates.}
To accommodate the unpredictability of mobile scheduling, we adopt a three-stage optimization schedule with progressively increasing trainable capacity. Let $\Theta$ denote all parameters, $\Theta_{\mathrm{head}}$ non-encoder parameters (heads and adapters), and $\Theta_{\mathrm{top}K}$ the top-$K$ Transformer blocks (counted from the output side):
\begin{equation}
\Theta_{1} = \Theta_{\mathrm{head}}, \quad
\Theta_{2} = \Theta_{\mathrm{head}} \cup \Theta_{\mathrm{top}K}, \quad
\Theta_{3} = \Theta.
\end{equation}
Stage~1 adapts only heads; Stage~2 adds partial encoder layers; Stage~3 allows brief full-model consolidation at a low learning rate. Training may stop after any stage if the device budget is exhausted.

\paragraph{Triplet discovery.}
Extracted entities are initially categorized into coarse types. For unseen entities, we extract stable semantic vectors via a frozen lightweight embedder and store them in a local vector database. When the count exceeds a threshold, agglomerative clustering with Ward linkage~\citep{ward1963hierarchical} groups semantically similar mentions. For clusters $A,B$ with sizes $|A|,|B|$ and embedding centroids $\boldsymbol{\mu}_A,\boldsymbol{\mu}_B$,
\begin{equation}
d(A,B) = \frac{|A|\cdot|B|}{|A|+|B|}\,\lVert\boldsymbol{\mu}_A - \boldsymbol{\mu}_B\rVert_2^2.
\end{equation}
Given threshold $\epsilon>0$, we iteratively merge clusters: whenever some pair $(A,B)$ satisfies $d(A,B)<\epsilon$, merge them; stop when no such pair remains (all surviving clusters are pairwise at least $\epsilon$-separated under $d$). The Thinking module then assigns category labels to large coherent clusters. For known-type entities, a lightweight relation classifier identifies relations, producing triplets written into MRGraph.

\subsection{Memory: MRGraph}
\label{sec:memory}

MRGraph is a three-layer on-device knowledge graph designed for continual writes and efficient retrieval with explicit provenance.

\textbf{Layer~1: Entity--paragraph graph.} The paragraph is the atomic unit for both writing and retrieval. Each paragraph node stores raw content, an in-document index, a pointer to the source document, and an embedding vector. Entity nodes are deduplicated via stable name hashing and linked to paragraphs through association edges. Relations are stored as keyed edges with observation counts to prevent uncontrolled graph growth. Ingestion follows an \emph{evidence-first, structure-second} principle: extracted triplets are first aligned to their best-matching paragraphs, then materialized as relation edges, ensuring every structured fact remains traceable.

\textbf{Layer~2: Abstraction layer.} Above entities, we maintain semantic clusters whose centroids are updated with smoothing as new entities arrive. An anchor--centroid drift metric detects and prevents topic drift. Cluster summaries are maintained incrementally during normal operation, with higher-quality refinement triggered during idle-and-charging periods.

\textbf{Layer~3: Runtime raw-text view.} At query time, retrieved paragraph evidence and the local graph neighborhood are assembled into a context bounded by a token budget, providing the Thinking module with both structured and unstructured evidence.

\subsection{Focus: hybrid retrieval}
\label{sec:focus}

At query time, the Thinking module decomposes the question into a retrieval plan $\mathcal{P} = (e_0, \{r_h, t_h\}_{h=1}^{H})$, where $e_0$ is the seed entity, $H$ is the number of hops, and for each hop $h\in\{1,\ldots,H\}$, $r_h$ is a natural-language relation phrase (e.g., ``\emph{published in}'') and $t_h$ is the desired target entity type for that hop.

\textbf{Stage~1: Plan-guided multi-hop text retrieval.} For each hop, the system constructs local queries by combining the current center entity with the relation phrase, retrieves candidate paragraphs using fused lexical and dense similarity scores, and selects evidence under a token budget via maximal marginal relevance (MMR). Retrieved paragraphs are used to infer next-hop entity centers.

\textbf{Stage~2: Graph-structured augmentation.} Bridge entities from Stage~1 seed a graph traversal. For each node, we compare its concatenation with the query against abstraction-layer summaries, select top-ranked clusters, and perform top-down retrieval to expand the subgraph. Personalized PageRank~\citep{page1999pagerank} estimates node importance, and connected subgraph blocks are assembled with supporting paragraphs.

\textbf{Stage~3: Full-document fallback.} If the first two stages provide insufficient evidence, global retrieval over documents provides a last-resort mechanism. Because on-device context length is limited, entities inevitably split across paragraphs; full-document fallback recovers these cross-boundary references.

Finally, retrieved evidence is post-processed: local context windows around entity mentions are extracted, sentence-level compression removes redundancy, and hash-based deduplication filters near-duplicates before packing the final context for the LLM.

\subsection{Thinking: LLM-driven semantic operations}
\label{sec:thinking}

The on-device LLM is treated as a reusable semantic engine invoked only for operations requiring high-level language understanding, through four roles:

\textbf{Labeling.} For entity clusters that cannot be confidently typed, the LLM proposes new type names. Let $\mathcal{M}$ denote the multiset of mentions awaiting labels; $\text{cluster}(\mathcal{M})$ groups them into clusters, and
\begin{equation}
\Delta\mathcal{T} = \text{LLM}\big(\text{cluster}(\mathcal{M})\big)
\end{equation}
returns a set of new type strings $\Delta\mathcal{T}$ for coherent clusters. Semantically dispersed clusters are rejected to avoid polluting the type space.

\textbf{Summarization.} The knowledge graph is periodically clustered into communities, and the LLM produces summaries re-inserted as upper-layer paragraphs, forming a self-compressing memory loop.

\textbf{Planning.} Complex questions are decomposed into multi-hop retrieval plans before evidence acquisition, reducing search ambiguity under strict on-device budgets.

\textbf{Answering.} The LLM generates responses conditioned on retrieved evidence, constrained to structured schemas that favor extractive grounding to reduce hallucination.

% ======================================================================
\section{Experiments}
\label{sec:experiments}
% ======================================================================

\subsection{Setup}
\label{sec:setup}
We use two compact LLM families in this work: Qwen3~\citep{qwen3report} and Ministral3~\citep{ministral3report}. Both are quantized to Q6\_K via \texttt{llama.cpp}~\citep{llamacpp2023}. All offline one-time preparation, including quantization, LoRA training, and dataset pre-processing, is performed on a cloud server with two NVIDIA RTX A6000 GPUs, while the full online SmartRAG pipeline---memory ingestion, indexing, retrieval, and answer generation---runs entirely on-device. For multi-hop retrieval planning, we train a task-specific LoRA decomposer~\citep{hu2022lora} on a self-constructed dataset derived from MultiHopQA and HotpotQA, where decomposition targets are generated offline using Claude Opus 4.6 and then filtered by rules with manual verification. The decomposer is used only to generate retrieval plans; final answers are always produced by the quantized backbone conditioned on retrieved evidence. All on-device experiments are conducted on two real smartphones: OnePlus~13 (Snapdragon~8 Elite, 16\,GB RAM) and OnePlus~15 (Snapdragon~8 Elite Gen~5, 16\,GB RAM).

We evaluate on four QA benchmarks: TriviaQA~\citep{joshi2017triviaqa}, Natural Questions~\citep{kwiatkowski2019natural}, HotpotQA~\citep{yang2018hotpotqa}, and MultiHopQA~\citep{ho2020constructing}. Baselines are divided into two groups. For \textbf{LLM-only} comparison, we evaluate generation without retrieval using Qwen3-1.7B, Qwen3-4B, Qwen3-32B, Ministral3-3B, and Ministral3-14B. Among these, Qwen3-1.7B, Qwen3-4B, and Ministral3-3B represent the compact-model regime that is executable on-device, with 1.7B/3B corresponding to practically deployable scales and 4B serving as a larger but still runnable case. By contrast, Qwen3-32B and Ministral3-14B are treated as large-model capacity references, allowing us to contextualize how far SmartRAG extends the capability of compact backbones. For \textbf{RAG}, we evaluate Na\"ive RAG~\citep{lewis2020retrieval}, InstructRAG~\citep{wei2024instructrag}, StableRAG~\citep{zhang2026stable_rag}, and TruthfulRAG~\citep{liu2026truthfulrag}, instantiated with two lightweight backbones, Qwen3-1.7B and Ministral3-3B, to match the compact-model regime targeted in this work. These baselines are compared under the same inference budget and retrieval context constraints; additional details on their cloud-side simulation setting are provided in \cref{sec:appendix_baselines}.

We report two complementary metrics. \textbf{Correctness (Cr)} measures whether a prediction is factually correct with respect to the reference answer, using an LLM-as-judger protocol~\citep{zheng2024judging} under a fixed rubric. \textbf{Token-F1} measures token-level overlap between prediction and reference using a rule-based normalization and matching procedure. Together, these metrics capture both factual validity and surface-level answer overlap. Full evaluation details, including the judge prompt and the Token-F1 computation procedure, are provided in \cref{sec:appendix_judge}.

\subsection{Main results: structured memory improves small-backbone QA under on-device constraints}
\label{sec:main_results}

\begin{table*}[t]
    \centering
    \caption{Correctness (Cr, \%) and F1 (\%) on four QA benchmarks. SmartRAG uses a quantized Qwen3-1.7B backbone running entirely on-device. Models $\ge$8B are evaluated on A6000 as capacity references. Within each base-model block, the best result in each column is shown in \textbf{bold} and the second-best in \underline{underlined}.}
    \label{tab:main}
    \small
    \renewcommand{\arraystretch}{1.1}
    \setlength{\tabcolsep}{3.8pt}
    \begin{tabular}{@{}ll cc cc cc cc@{}}
        \toprule
        \textbf{Base Model} & \textbf{Method} 
        & \multicolumn{2}{c}{\textbf{TriviaQA}} 
        & \multicolumn{2}{c}{\textbf{NQ}} 
        & \multicolumn{2}{c}{\textbf{HotpotQA}} 
        & \multicolumn{2}{c}{\textbf{MultiHopQA}} \\
        \cmidrule(lr){3-4}\cmidrule(lr){5-6}\cmidrule(lr){7-8}\cmidrule(lr){9-10}
        & & Cr & F1 & Cr & F1 & Cr & F1 & Cr & F1 \\
        \midrule

        \multicolumn{10}{@{}l}{\textbf{Qwen3}} \\
        \multicolumn{10}{@{}l}{\quad\textit{LLM-only (no retrieval)}} \\
        \quad & Qwen3-1.7B     & 22.10 & 21.52 & 9.10  & 10.17 & 14.15 & 13.00 & 22.00 & 19.18 \\
        \quad & Qwen3-4B       & 28.70 & 37.66 & 23.15 & 19.95 & 24.00 & 22.29 & 30.05 & 29.29 \\
        \quad & Qwen3-32B      & \textbf{51.45} & \underline{42.26} & 40.15 & 35.92 & 33.25 & 30.56 & 31.54 & 32.69 \\
        \addlinespace[2pt]
        \multicolumn{10}{@{}l}{\quad\textit{Retrieval-augmented (1.7B)}} \\
        \quad & Na\"ive RAG     & 47.00 & 36.31 & \underline{62.24} & \underline{37.23} & \underline{41.55} & \underline{37.80} & \underline{45.66} & \underline{34.58} \\
        \quad & InstructRAG     & 43.15 & 22.67 & 41.30 & 34.99 & 33.90 & 26.15 & 13.90 & 12.94 \\
        \quad & StableRAG       & 45.16 & 21.89 & 17.30 & 11.14 & 41.15 & 26.11 & 5.05 & 0.75 \\
        \quad & TruthfulRAG     & 47.10 & 22.86 & 45.10 & 26.38 & 40.71 & 25.51 & 19.95 & 12.64 \\
        \quad & \textbf{SmartRAG (ours)} & \underline{50.00} & \textbf{48.68} & \textbf{66.68} & \textbf{54.84} & \textbf{63.93} & \textbf{51.27} & \textbf{50.17} & \textbf{41.86} \\
        \midrule

        \multicolumn{10}{@{}l}{\textbf{Ministral3}} \\
        \multicolumn{10}{@{}l}{\quad\textit{LLM-only (no retrieval)}} \\
        \quad & Ministral3-3B   & 35.84 & 28.27 & 24.43 & 18.81 & 16.98 & 19.19 & 17.33 & 26.56 \\
        \quad & Ministral3-14B  & \textbf{60.70} & \textbf{60.87} & 37.80 & 34.55 & 22.40 & 25.34 & 20.12 & 31.45 \\
        \addlinespace[2pt]
        \multicolumn{10}{@{}l}{\quad\textit{Retrieval-augmented (3B)}} \\
        \quad & Na\"ive RAG     & 41.71 & 33.02 & 42.99 & 32.02 & 36.96 & 27.08 & 15.95 & 14.81 \\
        \quad & InstructRAG     & 50.25 & 28.74 & 50.60 & \underline{45.39} & 42.65 & \underline{37.41} & 28.75 & 26.27 \\
        \quad & StableRAG       & 15.95 & 10.86 & 27.35 & 18.11 & 44.03 & 32.94 & 13.71 & 9.07 \\
        \quad & TruthfulRAG     & 47.86 & 25.52 & \underline{53.00} & 31.31 & \underline{47.86} & 35.38 & \underline{43.20} & \underline{33.67} \\
        \quad & \textbf{SmartRAG (ours)} & \underline{59.74} & \underline{36.76} & \textbf{63.86} & \textbf{54.07} & \textbf{66.74} & \textbf{46.08} & \textbf{57.60} & \textbf{43.37} \\
        \midrule
        \bottomrule
    \end{tabular}
\end{table*}

\Cref{tab:main} compares SmartRAG against both LLM-only capacity references and lightweight retrieval baselines across the Qwen3 and Ministral3 families. Overall, SmartRAG delivers the clearest gains on NQ, HotpotQA, and MultiHopQA, while remaining competitive on TriviaQA. This pattern is consistent across both model families, suggesting that the main benefit of SmartRAG lies in structured evidence acquisition and composition.

\paragraph{SmartRAG is most effective on knowledge-intensive and evidence-composition benchmarks.}
Across both backbone families, the clearest gains of SmartRAG appear on NQ, HotpotQA, and MultiHopQA, where it consistently ranks at or near the top within each block. By contrast, TriviaQA is more mixed: SmartRAG remains competitive, but its advantage is less uniform than on the other three benchmarks. This pattern suggests that SmartRAG is especially effective when answering requires retrieving, organizing, and composing dispersed evidence, whereas direct open-domain factoid questions can often be handled well by larger parametric models or simpler retrieval strategies.

\paragraph{Comparison with RAG baselines.}
The baseline pattern also reflects the different objectives of the compared methods. InstructRAG explicitly learns denoising through self-synthesized rationales, StableRAG mitigates permutation sensitivity by aggregating reasoning across multiple retrieval orders, and TruthfulRAG focuses on resolving factual conflicts through triple extraction, KG retrieval, and conflict-aware filtering. These are all meaningful objectives, but they place more burden on the generator's ability to denoise, aggregate, or filter evidence. Under the smaller Qwen3-1.7B regime, such additional reasoning or filtering stages may not pay off as reliably as direct flat retrieval, which helps explain why these baselines often underperform Na\"ive RAG there. Under Ministral3-3B, however, the stronger compact backbone appears better able to benefit from these extra mechanisms, so the same baselines become consistently stronger than Na\"ive RAG. SmartRAG differs in that it does not rely primarily on one-shot denoising or conflict filtering; instead, it combines persistent structured memory with hybrid retrieval, which appears to be more robust across both compact backbone families.

\paragraph{Compact backbones versus larger capacity references.}
The goal of SmartRAG is to improve what compact backbones can achieve under strict on-device constraints. \Cref{tab:main} reflects this trade-off clearly. Larger standalone models still remain strongest on some direct factoid settings, most notably Qwen3-32B and Ministral3-14B on TriviaQA. At the same time, SmartRAG repeatedly matches or exceeds these larger LLM-only references on NQ, HotpotQA, and MultiHopQA. In conclusion, these results support the central claim of the paper: under device constraints, improving memory construction and evidence organization can be at least as important as increasing the parameter count alone.

\subsection{On-device feasibility}
\label{sec:device}

A central question for SmartRAG is whether the \emph{entire} pipeline remains usable when memory construction, retrieval, and answer generation are all executed on-device. To answer this, we profile two smartphone platforms with two different model sizes, and separately measure the end-to-end RAG pipeline. \cref{tab:device_configs} defines shorthand labels S0--S3 for each hardware--backbone pair, which are used throughout this section.

\begin{table}[t]
    \centering
    \small
    \caption{Device--backbone combinations for on-device profiling. Each pair is denoted \textbf{S0}--\textbf{S3}.}
    \label{tab:device_configs}
    \setlength{\tabcolsep}{8pt}
    \begin{tabular}{@{}l cc@{}}
        \toprule
        & \textbf{Qwen3-1.7B} & \textbf{Qwen3-4B} \\
        \midrule
        OnePlus~15 (Snapdragon 8 Elite Gen~5, 16\,GB) & S0 & S1 \\
        OnePlus~13 (Snapdragon 8 Elite, 16\,GB)       & S2 & S3 \\
        \bottomrule
    \end{tabular}
\end{table}

In general, the measurements reveal a clear usability boundary. While both backbones can be executed on-device, the 1.7B model remains in the practically deployable regime, whereas the 4B model substantially weakens interactivity once the full retrieval-augmented pipeline is taken into account.

We first examine resource usage during triplet extraction and QA. As summarized in \cref{tab:avg_memory_power}, the 4B backbone consistently incurs a substantially heavier memory footprint than 1.7B on both devices. This gap appears in both triplet extraction and end-to-end QA, where the 1.7B settings remain around the 4--5.5\,GB range while the 4B settings rise to roughly 6.5--6.8\,GB. This difference matters because SmartRAG runs as a full on-device pipeline, with model weights, retrieval-conditioned prefilling, and runtime memory operations sharing the same limited mobile memory budget.

More broadly, the measured power remains relatively low because we intentionally adopt a conservative system configuration: we prioritize stable long-term execution so that SmartRAG can remain deployable as a persistent on-device system. This choice leaves some hardware capacity unused, which in turn contributes to the non-trivial TTFT observed in the full QA setting.

\begin{table}[t]
    \centering
    \small
    \caption{Average PSS memory and power consumption across device--backbone configurations. S0--S3 follow the mapping in Table~\ref{tab:device_configs}.}
    \label{tab:avg_memory_power}
    \setlength{\tabcolsep}{6pt}
    \begin{tabular}{lcccc}
        \toprule
        \multirow{2}{*}{\textbf{Config}} & \multicolumn{2}{c}{\textbf{Triplet Extraction}} & \multicolumn{2}{c}{\textbf{QA}} \\
        \cmidrule(lr){2-3} \cmidrule(lr){4-5}
        & \textbf{PSS Memory (MB)} & \textbf{Power (W)} & \textbf{PSS Memory (MB)} & \textbf{Power (W)} \\
        \midrule
        \textbf{S0} & 5527.31 & 4.74 & 4149.72 & 2.85 \\
        \textbf{S1} & 6717.55 & 4.77 & 6803.68 & 5.36 \\
        \textbf{S2} & 5099.07 & 2.93 & 4153.38 & 2.12 \\
        \textbf{S3} & 6643.01 & 2.37 & 6512.62 & 2.12 \\
        \bottomrule
    \end{tabular}
\end{table}

\begin{figure}[t]
\centering
\includegraphics[width=0.75\textwidth]{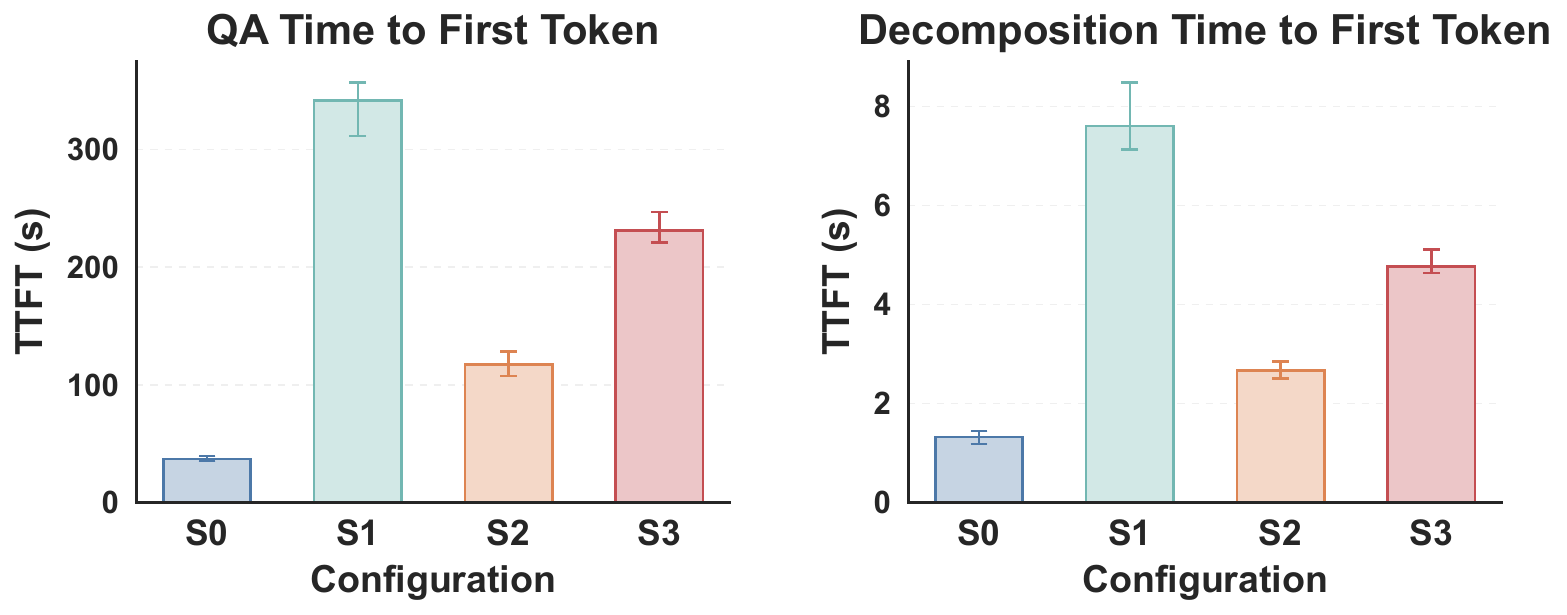}
\caption{Time to first token (TTFT) under configurations S0--S3. Left: QA TTFT. Right: question decomposition TTFT. Bars show median values and error bars indicate interquartile range.}
\label{fig:ttft_1x2}
\end{figure}

The more decisive separation appears in latency, shown in \cref{fig:ttft_1x2}. Question decomposition remains comparatively light-weight across all settings. For the S0 configuration (Snapdragon 8 Elite Gen~5 with Qwen3-1.7B), the median decomposition TTFT is \SI{1.32}{s}, which confirms that this planning stage remains near the low-second regime. By contrast, full QA is much more expensive because the model must absorb retrieved evidence before decoding the answer. Under the same S0 setting, the median QA TTFT rises to \SI{36.9}{s}, indicating that the dominant latency cost comes from retrieval-conditioned prefilling rather than from the decomposition step itself. The 4B settings (S1 and S3) are visibly slower in \cref{fig:ttft_1x2}, moving further away from an interactive operating point.

\begin{figure}[t]
\centering
\includegraphics[width=0.75\textwidth]{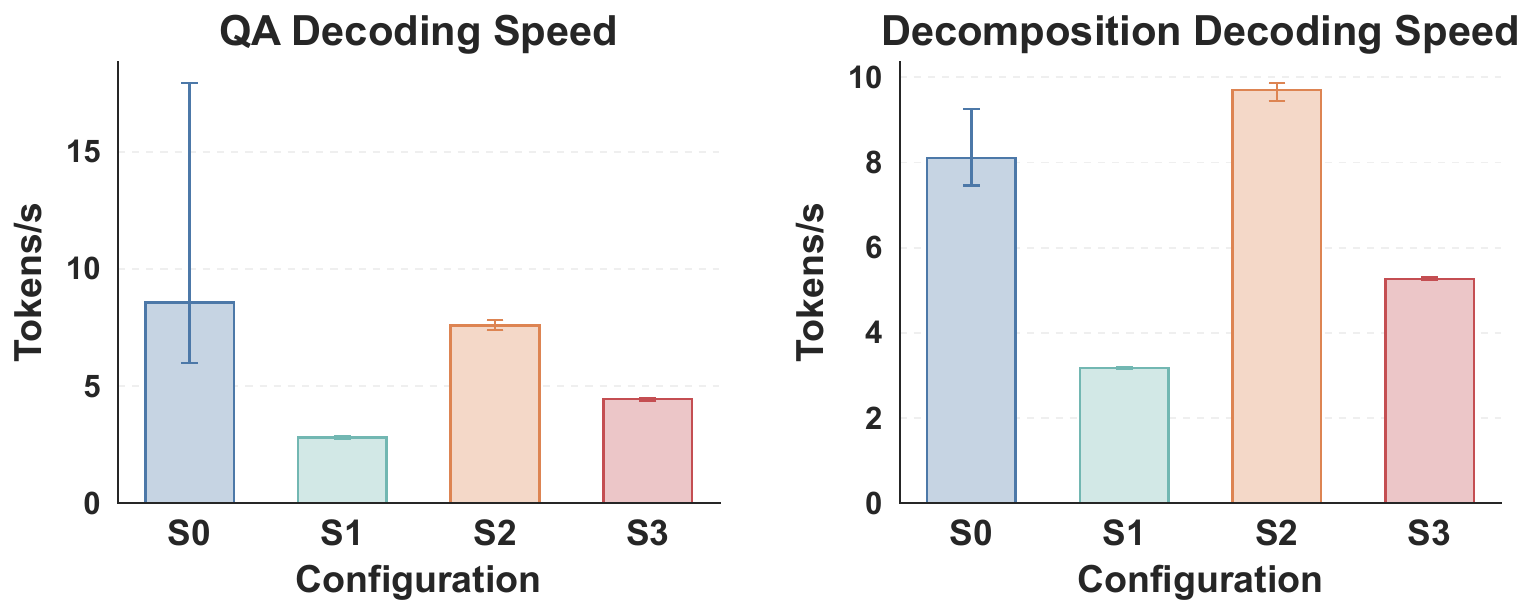}
\caption{Decoding speed under configurations S0--S3. Left: QA decoding speed. Right: question decomposition decoding speed. Bars show median values and error bars indicate interquartile range.}
\label{fig:decoding_speed_1x2}
\end{figure}

The same conclusion is reinforced by the generation stage in \cref{fig:decoding_speed_1x2}. In the S0 runlog, the median decoding time is \SI{8.11}{s} for question decomposition and \SI{8.55}{s} for QA, while the mean QA decoding time is higher (\SI{14.81}{s}) because answer lengths vary substantially across samples. This pattern again suggests that the main systems bottleneck is not the decomposition step itself, nor even decoding alone, but the long prefilling stage required by retrieval-conditioned QA. More broadly, \cref{fig:decoding_speed_1x2} shows the same qualitative ordering across devices and backbone sizes: the 1.7B backbone decodes faster than 4B, and the smaller configurations remain more favorable for end-to-end deployment.

The memory, TTFT, and decoding measurements explain why we standardize on the 1.7B backbone in the main accuracy experiments. Although the 4B model is technically executable on smartphones, its larger memory footprint, much longer retrieval-conditioned prefilling time, and slower generation place it beyond a practically interactive regime for end-to-end on-device RAG. We also note that system software still matters: even under closely related Snapdragon-class hardware, OS scheduling and runtime policies can noticeably affect realized throughput. The reported numbers should therefore be interpreted as conservative end-to-end baselines for deployable on-device operation rather than absolute hardware limits.

\subsection{Ablation study}
\label{sec:ablation}

\begin{figure}[t]
    \centering
    \includegraphics[width=0.45\linewidth]{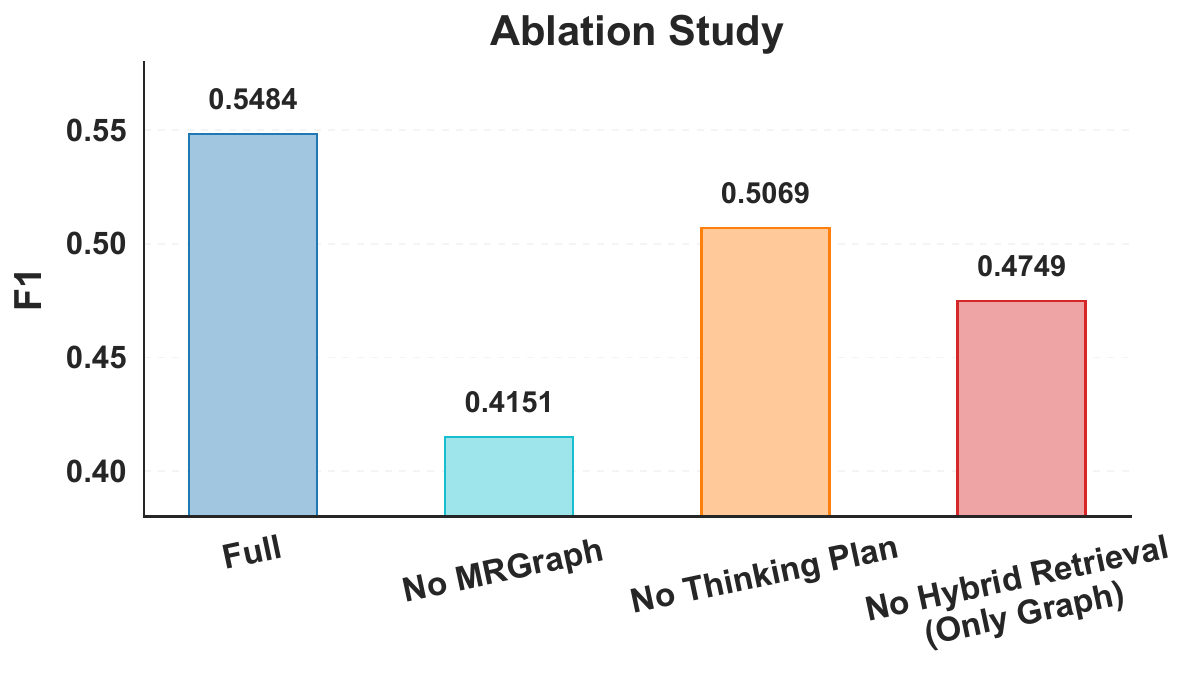}
    \caption{Ablation results on NQ. We compare the full SmartRAG system with variants removing MRGraph, the Thinking planner, or the hybrid retrieval design. Token-F1 is reported as the primary downstream QA metric.}
    \label{fig:ablation}
\end{figure}

Using MultiHopQA as the test set, we ablate three core components of SmartRAG: MRGraph (\texttt{no\_MRGraph}), the LLM-based retrieval planner (\texttt{no\_THINKING\_PLAN}), and the hybrid retrieval design (\texttt{no\_Hybrid\_Retrieval}, i.e., graph-only retrieval). The results are shown in \cref{fig:ablation}.

\paragraph{Key findings.}
As shown in \cref{fig:ablation}, the full SmartRAG system achieves the best performance, with a Token-F1 of 0.5484. Removing MRGraph causes the largest degradation, dropping Token-F1 to 0.4151, which indicates that the structured memory module is the most important component for downstream multi-hop QA in this setting. Disabling the Thinking planner also reduces performance to 0.5069, suggesting that explicit multi-hop planning improves evidence acquisition even when the remaining retrieval and answering pipeline is preserved. Replacing hybrid retrieval with graph-only retrieval further lowers Token-F1 to 0.4749, showing that graph traversal alone is insufficient and that SmartRAG benefits from combining graph-based signals with complementary retrieval channels. Overall, the ablation results support the full SmartRAG design: structured memory, planning, and hybrid retrieval each make non-trivial contributions, and their combination yields the strongest end-to-end QA performance.

% ======================================================================
\section{Discussion and conclusion}
\label{sec:discussion}
% ======================================================================

We have presented SmartRAG, an on-device architecture for adaptive question answering that decomposes intelligence into four coordinated functional roles. By shifting continuous adaptation to efficient structured computation and reserving LLM inference for high-value semantic operations, SmartRAG achieves multi-hop reasoning performance competitive with substantially larger cloud-based models while running entirely on commodity smartphones.

\paragraph{The retrieval--precision trade-off.}
Our analysis of the HotpotQA gap between SmartRAG and Na\"ive RAG reveals a fundamental trade-off in graph-structured retrieval. While graph traversal introduces relationships between nodes that enhance multi-hop reasoning, it also surfaces structurally related but contextually peripheral evidence. For open-ended questions where direct vector matching suffices, this can decrease recall. Intelligent search strategies that dynamically balance graph depth against retrieval precision represent a promising research direction.

\paragraph{Why not LLM-native extraction?}
Although a fully LLM-native extraction pipeline could reduce modular error accumulation and support richer predicate structures, our profiling shows that even query-time RAG prefilling can push TTFT to the minute scale on commodity smartphones. Making LLMs the primary writer of a continuously updated knowledge graph would introduce unacceptable latency and energy costs, and would require on-device training capabilities that remain impractical. The separation into a lightweight, trainable perception layer and a sparse LLM control plane yields predictable ingestion costs while retaining LLM expressivity where it matters most.

\paragraph{Limitations.}
(1)~The current planner is heuristic and does not uniformly benefit all question types; learned planning strategies could address this. (2)~On-device QA latency, while acceptable for the 1.7B backbone, is dominated by retrieval-conditioned prefilling and would benefit from evidence compression techniques. (3)~Typed NER remains low in AI and Politics despite surface gains; improving type-consistent discovery and clustering is still a major lever, even when Token-F1 and Hit@1 already move in the right direction. (4)~Our evaluation is limited to English-language benchmarks; multilingual adaptation remains future work.

\paragraph{Broader impact.}
SmartRAG enables privacy-preserving personal assistants that operate entirely on-device without sending user data to the cloud. While this is broadly positive for user privacy, we note that any system capable of building detailed personal knowledge graphs raises data governance concerns if the device is compromised. Appropriate encryption and access control mechanisms should complement the on-device architecture.

% ======================================================================
\begin{ack}
Acknowledgments will be added in the camera-ready version.
\end{ack}
% ======================================================================

{\small
\bibliographystyle{plainnat}
\bibliography{references}
}

% ======================================================================
\appendix
% ======================================================================

\section{LLM-as-judger protocol}
\label{sec:appendix_judge}

We assess factual correctness using an LLM-as-judger approach~\citep{zheng2024judging}. Given a question, a reference answer (or set of acceptable answers), and the model's prediction, a judge LLM (Claude Haiku~4.5) determines whether the prediction is factually correct under a strict rubric. The judge outputs a binary correctness label per sample; we report the mean across all samples as \textbf{Correctness (Cr, \%)}. SQuAD-style \textbf{F1} is computed as standard token-level precision--recall against reference answers. All models are evaluated under identical inference budgets: 4096 token context, top-$K$=8 retrieval cutoff, and Q6\_K quantization.

\section{Device specifications}
\label{sec:appendix_devices}

\begin{table}[h]
    \centering
    \small
    \caption{Evaluation device specifications.}
    \begin{tabular}{@{}lll@{}}
        \toprule
        & \textbf{OnePlus 13} & \textbf{OnePlus 15} \\
        \midrule
        SoC        & Snapdragon 8 Elite    & Snapdragon 8 Elite Gen 5 \\
        RAM        & 16\,GB LPDDR5X         & 16\,GB LPDDR5X \\
        Storage    & 256\,GB UFS 4.0        & 256\,GB UFS 4.0 \\
        OS         & ColorOS 16             & ColorOS 16 \\
        \bottomrule
    \end{tabular}
\end{table}
\section{Baseline implementation details and deployment notes}
\label{sec:appendix_baselines}

To make the comparison with prior RAG baselines as fair and transparent as possible, we implemented Na\"ive RAG, InstructRAG, StableRAG, and TruthfulRAG under a unified evaluation backend and quantization setting. In all cases, we used the same \texttt{llama.cpp} inference backend and the same model quantization scheme as in our main experiments, and instantiated the baselines with the same compact backbone families reported in \cref{tab:main}. This appendix clarifies an important distinction: unlike SmartRAG, these baselines were \emph{not} profiled as end-to-end real-phone deployments. Instead, they were executed in a cloud-side simulation environment designed to match the same inference backend, quantized models, and retrieval budget as closely as possible.

\paragraph{Why these baselines were not deployed as real-phone end-to-end systems.}
Our goal in the baseline comparison is to evaluate answer quality under a matched compact-backbone regime, rather than to claim that every baseline is equally suitable for direct mobile deployment. The original formulations of InstructRAG, StableRAG, and TruthfulRAG introduce additional stages that make them difficult to treat as drop-in phone pipelines. InstructRAG explicitly denoises retrieved contents through self-synthesized rationales and supports both in-context learning (ICL) and supervised fine-tuning (SFT). TruthfulRAG introduces additional KG-oriented processing steps, including knowledge-graph construction, graph retrieval, and entropy-based filtering. StableRAG emphasizes repeatability, frozen retrieval, and structured output normalization, with a pipeline that depends on OpenAI-style generation and embedding calls together with a persistent retrieval store. These original designs are meaningful in their intended settings, but they differ substantially from a single-pass compact on-device assistant stack. For this reason, we treat them as \emph{reference RAG baselines evaluated under a matched backend and quantization setup}, rather than as directly deployable mobile systems.
\paragraph{InstructRAG.}
InstructRAG is a framework that explicitly denoises retrieved contents by generating rationales, and supports both in-context learning (ICL) and supervised fine-tuning (SFT). In our experiments, we implemented the method as described in the original paper, replacing the original serving stack with the same quantized compact backbones and \texttt{llama.cpp} backend used in our main experiments. We did not attempt to reproduce the original large-scale fine-tuning regime; instead, we adapted the method to the same compact-backbone comparison regime used for the other baselines. We therefore interpret the resulting numbers as measuring how well the \emph{InstructRAG-style denoising strategy} transfers to compact quantized backbones under a unified inference setting, rather than as a claim about the maximum achievable performance of the original method.
\paragraph{TruthfulRAG.}
TruthfulRAG is a KG-based framework for resolving factual-level conflicts in RAG, whose pipeline includes knowledge-graph generation, graph retrieval, and entropy-based filtering before final response generation. In our evaluation, we preserved this overall conflict-aware graph-processing logic while replacing the original model-serving setup with the same compact quantized backbones and \texttt{llama.cpp} backend used elsewhere in the paper. Because TruthfulRAG contains additional graph-construction and filtering stages beyond a standard lightweight RAG loop, we did not deploy it as an end-to-end real-phone pipeline. Instead, we evaluate it in the same cloud-side simulation environment as the other RAG baselines. This makes the comparison more controlled at the backbone and inference level, while acknowledging that TruthfulRAG was originally designed for a somewhat different operating point than a strict mobile deployment target.
\paragraph{StableRAG.}
StableRAG is a method designed to improve repeatability and robustness in RAG systems, mitigating retrieval-permutation-induced hallucinations through stable retrieval ordering and structured rendering. In our experiments, we retained the stability-oriented evaluation logic described in the original paper while replacing the original serving configuration with the same compact quantized backbones and \texttt{llama.cpp} backend used in the rest of the paper. As with TruthfulRAG and InstructRAG, we treat the resulting system as a \emph{simulated reference implementation under matched inference conditions}, not as a true smartphone deployment. This distinction is important because StableRAG is primarily designed to study response stability and retrieval repeatability, whereas SmartRAG is optimized for persistent structured memory and end-to-end on-device execution.
\paragraph{Interpretation of the comparison.}
We emphasize that these baselines are included as strong and relevant points of comparison, not as strawmen. InstructRAG targets denoising under noisy retrieval; TruthfulRAG targets factual conflict resolution with KG support; and StableRAG targets repeatability and robustness to retrieval variation. These are all reasonable objectives, and each may be better aligned than SmartRAG with other deployment scenarios. Our comparison should therefore be read narrowly: under a unified compact-backbone, quantized, \texttt{llama.cpp}-based setting, SmartRAG is more effective for the particular combination of goals studied in this paper---namely, persistent structured memory, repeated evidence composition, and end-to-end QA under edge-style resource constraints. We do not claim that the released original forms of these methods are inferior in general, nor that they were designed for the same deployment target as SmartRAG.

\end{document}